# A rule based algorithm for detecting negative words in Persian

**Reza Takhshid**
Reza.takhshid95@student.sharif.edu
**Adel Rahimi**
Adel.rahimi@mehr.sharif.edu

In this paper, we present a novel method for detecting negative words in Persian. We first used an algorithm to an exceptions list which was later modified by hand. We then used the mentioned lists and a Persian polarity corpus in our rule based algorithm to detect negative words.

# 1. Introduction

Persian language is the official language of Iran. Its two varieties are the official languages in Afghanistan and Tajikistan. Persian which is called Farsi by the native speakers, due to it being closely related to Middle and Old Persian: the former language of Fars in southwestern Iran, is the native language of almost 62 million people.

In order to facilitate sentiment analysis of Persian text, we've designed and implemented an algorithm which aims to detect words with negative polarity. Currently most sentiment analysis algorithms depend mainly on polarity datasets. However, since negative prefixes in Persian are only attached to a semantically positive base (shaghagi, 2002), we have incorporated a negative prefix detection to further increase the functionality of sentiment analysis in Persian.

Other languages specially English have extensive sentiment databases. To name a few, Bradley (1999) created a list of words based on several psychological factors such as pleasure. Esuli (2007) made sentiwordnet, an extension of wordnet but with the sentiment of the words in it. As Liu (2012) suggests sentiment lexicon is necessary for sentiment analysis but it's not enough. Taboda (2011) suggests a lexicon based method for sentiment analysis, and Booster wordlist is a lexicon of words that "boost or reduce subsequent words" Thelwall (2010). Emoticon list is also a sentiment lexicon. Saleh (2011), and Urdu, Syed (2014).

In contemporary Persian seven negative prefixes are used to build words with negative or contrastive meaning (shaghaghi, 2002). These prefixes are: ن, لا, غیر, ضد, بی, نا, پاد.

## 2. Data sets

Two sets of data are used in our algorithm. The first is the "Polarity Corpus of Persian lexicon" (Dehdarbehbahani, 2014) developed in intelligent systems laboratory of University of Tehran. There are 961 negative words which are tagged by hand. The second is an exceptions list which we extracted using the Flexicon database and Bi Jan Khan Corpus of Persian language which contains a little over two and a half million word.

To create this exceptions list, first, we generated a raw list of all the words in the aforementioned corpus and database which began with any of prefixes mentioned before. Next, a monogram model of Bi Jan Khan corpus was built. After that, for every word in the raw list we removed the "prefix" and looked up the remaining part of the word in the monogram model. If the count was more than five, that word was added to a text file, let us call it ValidAffixed. On the other hand if "unprefixed" part appeared less than five times in the corpus it was added to another text file, we call this one Exceptions. Table1 shows the size of each file.

| Source | Word Count |
| --- | --- |
| Flexicon Exceptions | 2669 |
| Flexicon ValidAffixed | 1416 |
| Bi Jan Khan Exceptions | 2256 |
| Bi Jan Khan ValidAffixed | 1105 |

Table1. Raw data of the exception extraction phase

As the next step we checked these files by hand and corrected the mistakes made in the exception extraction phase. There are a few reasons as to why the exceptions and valid forms were wrongfully detected. The main reason would be that the base of some validly affixed forms do not appear frequently in the corpus, for example: بی‌بروبرگرد or بی‌چشم‌ورو. The same problem occurs with Arabic loan words: لاینفک or لاینقطع. Moreover sometime prefixes are attached to bound morphemes, and upon the failure of the base to appear more than five times in the corpus the words were added to the exceptions list. On the other hand some words like بیان or بیگودی were not added to the Exceptions file because آن and گودی freely appear in the corpus with high frequency. Finally there is also the problem of homographs like نِشسته and نَشسته which can only be fixed by having their phonological representation. In the end, we have a list with 4168 exceptions, the words which their "prefix" is actually a part of the word: لارستان or نارنگی. we've also included several words which are in fact validly prefixed however the polarity of the outcome is actually positive, some example are: ضدآب or پادزهر.

## 3. The algorithm

Every input is first preprocessed. our preprocess function first strips the input, then calls Hazm normalizer, after that replaces spaces with \u200c character. This causes no problems since our inputs are not sentences but tokens. And finally the input is stemed[1]. The stemmer is specially written to only remove postfixes from nouns.

Having the polarity data and the exception list, the algorithm itself is fairly simple. First we check the polarity data for a string matching the input word from its first character to the $n^{th}$, n being the length of the input which decrements until it's value is one. If a match is found the program returns True, meaning the input is negative. Next we search the input in the exceptions list, if it is found, the program returns False. Finally we well check the input against all negative prefixes if the word starts with any of the prefixes True is returned otherwise False is return. A flowchart of the algorithm follows:

## 4. the results

We ran our program on a list of 100 randomly chosen words from around then then. It returned 20 negative words out of witch 14 were actually negatives. And only 2 negatives were undetected in the non-negative list it return. Although a more suitable test would be to use the program in an actual sentiment analysis software and compare the results with the old negative detections.

---

[1]Stemmer writen by Reza Takhshid.

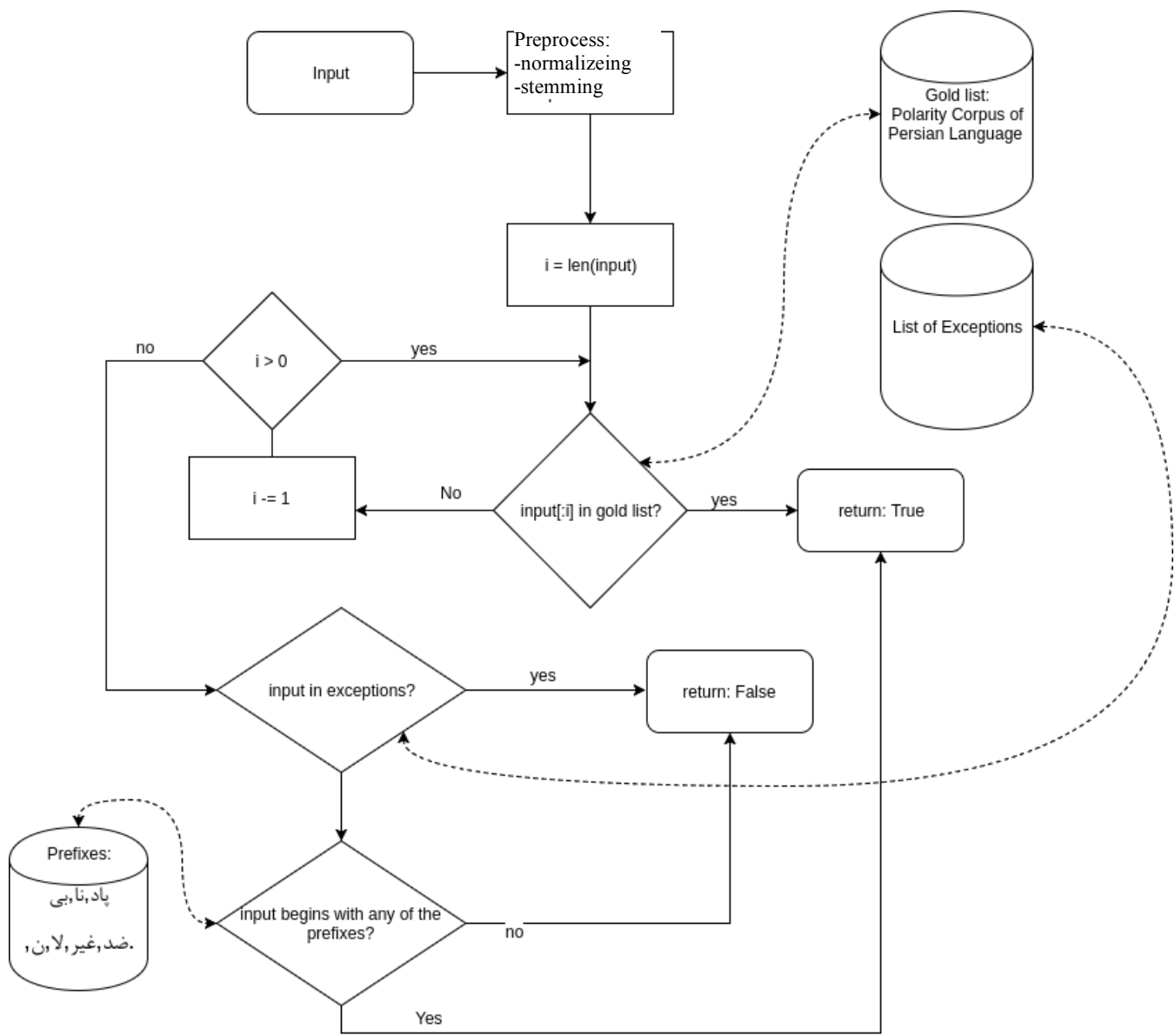

# 5. refrences